# Reviving Threshold-Moving: a Simple Plug-in Bagging Ensemble for Binary and Multiclass Imbalanced Data

Guillem Collell[1,2], Drazen Prelec[2,3,4], Kaustubh R. Patil[2,*]

[1]Computer Science Department
KU Leuven, Heverlee 3001, Belgium
[2]MIT Sloan Neuroeconomics Lab
[3]Departments of Economics and [4]Brain & Cognitive Sciences
Massachusetts Institute of Technology, Cambridge MA 02142
`gcollell@kuleuven.be, dprelec@mit.edu, kaustubh.patil@gmail.com`

*Abstract*. Class imbalance presents a major hurdle in the application of data mining methods. A common practice to deal with it is to create ensembles of classifiers that learn from resampled balanced data. For example, bagged decision trees combined with random undersampling (RUS) or the synthetic minority oversampling technique (SMOTE). However, most of the resampling methods entail asymmetric changes to the examples of different classes, which in turn can introduce its own biases in the model. Furthermore, those methods require a performance measure to be specified *a priori* before learning. An alternative is to use a so-called threshold-moving method that *a posteriori* changes the decision threshold of a model to counteract the imbalance, thus has a potential to adapt to the performance measure of interest. Surprisingly, little attention has been paid to the potential of combining bagging ensemble with threshold-moving. In this paper, we present probability thresholding bagging (PT-bagging), a versatile *plug-in* method that fills this gap. Contrary to usual rebalancing practice, our method preserves the *natural* class distribution of the data resulting in well calibrated posterior probabilities. We also extend the proposed method to handle multiclass data. The method is validated on binary and multiclass benchmark data sets. We perform analyses that provide insights into the proposed method.

## 1 Introduction

Dealing with class imbalance is an important but unsolved problem in data mining [1]. Imbalanced data sets frequently appear in real-world problems, for example, in fault and anomaly detection [2, 3], fraudulent phone call detection [4] and medical decision-making [5] to name a few. Standard learning algorithms are often guided by global error rates and hence may ignore the minority class instances leading to models biased towards predicting the majority class. Several methods have been proposed to alleviate this problem (see, e.g. [6, 7] for review). Often, a first choice to deal with the imbalance consists of pre-processing the data by resampling to balance class distributions [8, 9]. This is often achieved by either randomly oversampling (ROS) the minority class [10] or randomly undersampling (RUS) the majority class [11]. More informative methods that generate synthetic instances of the minority class are also a popular choice, e.g. the synthetic minority oversampling technique (SMOTE). For brevity we will henceforth call those data pre-processing methods collectively as rebalancing methods. Such data pre-processing are often combined with ensemble methods to leverage on their advantages over a single classifier [12]. Many of such combinations have been shown to be a powerful solution for imbalanced classification [6, 13, 14]. However, there are three major drawbacks of rebalancing methods; (1) potential loss of informative data (for undersampling), (2) changes in the properties of the data which in turn can cause the models to learn unwanted biases, e.g. miscalibrated posteriors [15, 16] and (3) it is often not evident which class distributions to use for a given data set and a performance measure of interest. Wrapper methods [17] can be employed but they are computationally expensive and often cater towards only a single objective, e.g. either accuracy or F1. For multiclass data, this problem becomes even more pronounced as there can be multiple minority or multiple majority classes. Making it is nontrivial to extend the balancing heuristics normally used for binary data to multiclass data [18].

Using decision thresholds is an alternative technique that can deal with class imbalance. The main difference between rebalancing and threshold-based methods lies in that the former relies on data pre-processing before the learning phase while the later on manipulating the model output. This technique has been utilized with some popular learning methods including with a small ensemble [19–21]. However, to our knowledge, the combination of threshold-moving with a bagging ensemble has not been thoroughly investigated. Bagging ensembles are a good candidate to combine with threshold-moving methods as they are known to provide good probability estimates [22].

In particular, we seek a method that provides well calibrated posterior probability estimates. An important advantage of such a method will be that it can be used as a *plug-in* method. A plug-in method is a versatile method such that the threshold can be set *a posteriori*, i.e. at the test phase [23]. This is a major improvement over cost-sensitive and rebalancing methods which have to set the correct loss at the learning phase. We propose PT-bagging that, as we will show, passes as a plug-in method. The main motivation behind PT-bagging is to leverage advantages of bagging while avoiding the problems that rebalancing methods inexorably entail. We show that the proposed method addresses those problems and possesses several desirable properties:

(1) It can be used as a plug-in method to maximize a performance measure of interest without retraining but by just applying a different threshold *a posteriori*. By contrast, rebalancing methods are not very flexible and need computationally expensive parameter tuning, e.g., which class distributions to use for learning via a wrapper approach [17].
(2) It consistently performs close to the best possible macro accuracy performance without the need of finding the optimal threshold empirically (e.g., by cross-validation). Note that having a validation set for fine-tuning can be computationally costly, it might not always possible or might be expensive (e.g., data collection).
(3) It is directly extendable to multiclass data.

In this respect, we provide a theoretical analysis on when optimal macro accuracy performance is guaranteed. By contrast, we empirically show that, in general, rebalancing methods could attain higher macro accuracy by choosing a different decision threshold. However, for other measures, such as the F1 score, it is not possible to obtain a closed expression for the optimal costs/thresholds [24]. Nevertheless, we show that with a simple and sensible threshold PT-bagging achieves better F1 performance than most of the evaluated methods. In this respect, the paper makes two additional contributions; (1) proposal of a threshold for maximizing the macro F1-score and (2) comparison and analysis of the "full potential" of the methods.

The rest of this paper is organized as follows. In section 2 we introduce the relevant background for our work. In the same section, we describe some popular resampling ensemble methods and we discuss their potential flaws. In section 3, we describe our proposed algorithm (PT-bagging) and provide a theoretical justification of its performance. In the following section, we describe our experimental setup. Next, we present a complete set of empirical tests in order to compare method performance and a discussion of the results. Finally, in conclusions and future work, we discuss the implications of our findings and propose future lines of research.

## 2 Background

We consider the classification problem setting where the aim is to learn from training data tuples $\{x_i, y_i\}_{i=1}^{N}$, where $x_i \in R^d$ and $y_i \in \{1, ..., m\}$. The learned model is then used to make predictions on new test data $\{x_j\}_{j=1}^{M}$. For binary data we have $y_i \in \{0,1\}$ and we denote the minority class as the class 1. We define the class specific thresholds as $\lambda_i, i = 1, ..., m$. We make two assumptions; (1) the probability distribution of the test data is similar to that of the training data and (2) the class distribution of the training data provides accurate estimate of their respective prior probabilities.

### 2.1 Performance measures in imbalanced learning

Accuracy (correct classification rate) is a good metric when data sets are balanced. However, it is a misleading measure in the imbalanced domain. For example, the naïve strategy of classifying all the examples into the majority class would obtain 99% accuracy in an imbalanced data set composed of 99% examples of the majority class. Therefore, other measures are used when the class distribution is imbalanced. A confusion matrix (Table 1) summarizes the performance of a classifier.

Table 1. Confusion matrix in binary classification

|  | Predicted positive (1) | Predicted negative (0) |
|---|---|---|
| Actual positive (1) | TP (true positive) | FN (false negative) |
| Actual negative (0) | FP (false positive) | TN (true negative) |

There exist several widely accepted performance measures in imbalanced learning, all of them computable from the elements of the confusion matrix. Some of the most extensively used measures are as follows:

$$\text{Recall (=TPR)} = \frac{TP}{TP+FN}; \quad \text{Precision} = \frac{TP}{TP+FP}$$

$$\text{Macro accuracy} = \frac{TPR + TNR}{2} \quad \text{F1-score} = \frac{2TP}{2TP+FP+FN}$$

where TNR=TN/(FP+TN). Macro F1-score is calculated by considering each class separately as the positive class and then averaging the corresponding F1-scores. In addition, the receiver operating characteristic (ROC) curve is often employed [6]. The ROC curve is generated by plotting TPR (y-axis) and FPR=FP/(FP+TN) (x-axis) at different decision thresholds. Hence, the point (0, 1) equals to a perfect prediction whilst the line $x = y$ corresponds to random guessing. We notice, however, that ROC curves suffer from a serious limitation for evaluating performance under class imbalance. When data is highly imbalanced, ROC curves fail to capture large changes in the number of false positives since the denominator of FPR is largely dominated by TN. Precision-recall curves (PR) are preferred over ROC curves for evaluating method performance in the imbalanced domain [7]. Therefore, in this work we use PR curves instead of ROC curves.

## 2.2 Learning from imbalanced data

Many solutions have been proposed to deal with the class imbalance problem in classification. These solutions mainly fall into one of the following three major strategies: (1) cost-sensitive learning; (2) resampling techniques, and (3) threshold-moving, each of which is briefly discussed below.

(1) Cost-sensitive learning places different misclassification costs on different classes. Higher misclassification costs for the minority class are imposed by a loss function in order to guide the learning process. In fact, by altering the training class distribution, resampling techniques effectively impose different misclassification costs [25]. For this reason, we only consider resampling methods in this work.

(2) Rebalancing techniques modify the class balance in the training data. Such data pre-processing solutions do not entail any modification in the learning algorithm and therefore can be readily employed with standard learning algorithms. They will be discussed later.

(3) Threshold-moving: a classifier is learnt in a data set with the original class proportions and its probabilistic prediction is "corrected" at test time according to the class priors. In binary classification, the threshold for predicting a given class is usually set equal to the class prior.

### 2.3 Bagging ensemble

Ensemble methods make use of many classifiers and they often show improvement over individual classifiers' performance [26]. Two popular ensemble techniques are boosting and bagging [27]. However, in this work we focus on bagging ensembles (Algorithm 1) since boosting ensembles are known not to perform better in the imbalanced domain [6].

---

**Algorithm 1.** Pseudo-code for bagging ensemble.

```
1. Learning:
1.1. Input: Training set S = {(xᵢ, yᵢ)}, i = 1, …, N; and yᵢ ∈ { 1, …, m}
where m is number of class labels and n the number of classi-
fiers.
1.2. Generate n training data sets by sampling S.
1.3. Learn n base classifiers from each sample.

2. Prediction:
2.1. Input: instance x, base classifiers, probability thresh-
old for each class.
2.2. Each base classifier i gives a probabilistic estimate
Pᵢ(y = k|x) for each label k ∈ { 1, …, m} given a test instance x.
2.3. Compute averages of probabilistic predictions for each
class k ∈ {1, …, m}:  P(y = k|x) = (1/n) Σᵢ Pᵢ(y = k|x).
2.4. Normalize the averaged probabilities to sum to 1.
2.5. Rank each class k ∈ { 1, …, m} according to: P(y = k|x)/λₖ.
2.6. Assign the label for which the score in 2.5 is highest.
```

---

In bagging, each classifier learns from a different sample of the original training set. The main principle behind bagging's performance is that the averaged prediction reduces variance of individual classifiers without increasing their bias. Bagging only performs well with unstable (high variance) base classifiers, i.e., classifiers such that small changes in the training data lead to large changes in the learned model [27]. For example, decision trees and neural networks are unstable classifiers and thus suitable to be combined with bagging. Clearly, different sampling mechanisms (Algorithm 1, step 1.2) and different priors (step 2.5) can be used which will lead to different models. We discuss some relevant sampling mechanisms in the next section.

At the prediction phase the individual predictions are aggregated. Different aggregation methods can be used, e.g. hard-voting that uses crisp class assignments or soft-voting that uses probabilistic predictions. It is known that soft-voting provides better performance than hard-voting [21, 28, 29]. As we will show the formulation in Algorithm 1 allows our method to be directly extended to multiclass problems, which is not trivial when the threshold is considered as a cut-off.

### 2.4 Resampling mechanisms

In this section we briefly describe the resampling mechanisms that can be used together with the bagging ensemble by adapting the step 1.2 in Algorithm 1. One of the simplest ways to resample is to sample each example with equal probability with replacement, i.e. non-parametric bootstrap. This sampling is not commonly used with imbalanced data as it preserves the imbalanced class distribution which is thought to be detrimental to learning. In this paper, we argue that this is actually a good mechanism to use.

Other sampling mechanisms used for imbalanced data try to rebalance the class distribution. Perhaps the simplest and most popular undersampling mechanism used for ensemble learning is the exactly balancing (EB). EB sampling preserves all the minority class instances while randomly undersampling majority class to exactly balance the class proportions. Roughly balancing (RB) is a powerful variation of EB that improves performance by increasing diversity of the classifiers [13]. RB, like EB, preserves the minority class examples but under-samples the majority class examples roughly balancing (as determined by a negative binomial distribution) the class distribution. RB generally outperforms EB when combined with bagging.

Random oversampling methods, on the other hand, over-sample the minority class examples. Results from previous studies indicate that undersampling generally performs better than oversampling [6, 30, 31]. Furthermore, undersampling is computationally efficient since a large part of the training data is discarded. More sophisticated oversampling methods have been proposed, one of the most popular ones is the SMOTE. SMOTE generates new minority class examples by interpolation, which is then combined with the undersampling of majority class examples. SMOTE often performs well in combination with bagging ensemble. Note that the common threshold of 0.5 is normally used with the rebalancing methods (Algorithm 1, step 2.5).

Taken together, an important effect of rebalancing is that it can lead to miscalibrated posterior probability estimates as recent studies have pointed out [15, 16]. An additional (and usually unnoticed) potential problem of resampling techniques is that of dataset shift, i.e. differing training (balanced) and test (natural class distribution) probability distributions [16]. This might carry some additional problems since a model that was optimized for balanced data is then evaluated in a different condition for which the optimality might no longer hold. By contrast, the simple bootstrap sampling does not modify the data distribution, which leads us to hypothesize that it will be less prone to those problems.

# 3 Probability threshold bagging (PT-bagging)

Several studies have considered thresholding as a method to deal with class imbalance [19, 32]. For example, Maloof (2003) [19] compared threshold-moving method to RUS, both applied to a single classifier, and concluded that they achieve similar performance in terms of ROC. However, to our knowledge, in none of the previous studies thresholding was studied in combination with bagging methods. It is a common practice, however, to use bagging ensembles with resampling mechanisms [6].

The basic idea behind PT-bagging is to leverage bagging (Algorithm 1) to get well calibrated posterior estimates and to appropriately threshold them afterwards, according to the performance measure to be maximized. PT-bagging learns base classifiers from simple bootstrap replicates of the original data set which preserve the class distribution. Then, following Algorithm 1, probabilistic predictions for each class k are averaged across individual classifiers, to obtain a final posterior probability estimate $P(y = k|x)$. To get a class label, the probability estimate is transformed into a score by dividing it by its respective class threshold $\lambda_k$. The class $k$ for which this ratio $P(y = k|x)/\lambda_k$ is the largest is assigned. According to the thresholding methods' categorization proposed by Hernandez-Orallo et al. [33], PT-bagging employs a so-called score-driven threshold, as opposed to e.g., a fixed threshold such as always using 0.5. For example, to maximize macro accuracy, the optimal threshold for class $k$ equals to the class prior, i.e., $\lambda_k = P(y = k)$ in the training data (see Theorem 1 below). For instance, if the normalized averaged probabilistic prediction of individual classifiers for class 0 is 0.7, and the proportion of this class is 0.8 (i.e., $P(y = 0) = 0.8$), then the score of class 0 is 0.7/0.8=0.875, consequently the score for class 1 is 0.3/0.2=1.5. Thus, in that case, class 1 will be predicted. The calculation for this threshold is identical in the multiclass setting. We denote the method with a threshold that maximizes a particular measure using a subscript; $PT_{MA}$-bagging for macro accuracy and $PT_{F1}$-bagging for macro F1-score.

## 3.1 Threshold for maximizing macro accuracy

The following theoretical results aims at providing insight on the mechanism behind the performance of thresholding for the macro accuracy measure. The reader should note that the main message of Theorem 1 is not the finding of the optimal misclassification threshold for the macro accuracy measure (which are known to be equal to the prior of the minority class), but a constructive proof of an optimal algorithm that maximizes such measure for binary and multiclass data. Theorem 1 shows that a necessary condition for a method to maximize macro accuracy is to have good estimates of the posterior probabilities $\widehat{P}(y = k|x)$ for each class $k$.

***Theorem 1***. Proposition: Let $P(y = j)$ be the prior on class $j$ and $P(y = j|x)$ the true (unknown) posterior probability of class $j$ given $x$. If proportions of each class are unchanged from training to test, then predicting class $k$ such that

$$k = \underset{j}{\mathrm{argmax}}\ \frac{P(y=j|x)}{P(y=j)} \qquad (1)$$

maximizes the macro accuracy

*Proof:*

To simplify notation and improve readability, let us consider the binary class problem. The same proof trivially generalizes to multiclass. Let $C = \{1,0\}$ be the class labels (positive=1 and negative=0). Let $D = k$ be a prediction of class $k$ given by classifier $D$. Thus, $P(D = k|x)$ is the probability for classifier $D$ to predict class $k$ at $x$.

We shall first derive a suitable expression of macro accuracy. Recall that macro accuracy equals to (TPR+TNR )/2 where TPR=TP/P and TNR $=$ TN/N. We remind the notation P=TP+FN and N $=$ TN $+$ FP. Let us first derive a continuous expression for TPR. Notice first that TP/(P $+$ N) $= \int_R P(y=1|x)P(D=1|x)p(x)dx$ and that P/(P $+$ N) $= P(y=1)$. Thus, the ratio of the first expression over the second one equals to TPR=TP/P. That is

$$TPR = \frac{\int_R P(y=1|x)P(D=1|x)p(x)dx}{P(y=1)}$$

The derivation of TNR is analogous. Thus, dividing the sum of TPR and TNR by 2, one obtains the expression of macro accuracy:

$$\frac{1}{2}\frac{\int_R P(y=1|x)P(D=1|x)p(x)dx}{P(y=1)} + \frac{1}{2}\frac{\int_R P(y=0|x)P(D=0|x)p(x)dx}{P(y=0)}$$

By entering both terms into the same integral:

$$\frac{1}{2}\int_R \{\frac{P(y=1|x)}{P(y=1)}P(D=1|x) + \frac{P(y=0|x)}{P(y=0)}P(D=0|x)\}p(x)dx \qquad (2)$$

Therefore, maximizing the integral in (2) is equivalent to asking for the optimal choice of $P(D = 1|x)$ and $P(D = 0|x)$ at each $x$. In other words, how to assign class labels 1 or 0 in a wise way (perhaps probabilistically), given $x$. Notice that the inside of the integral in (2) is nothing but the following convex combination

$$\frac{P(y=1|x)}{P(y=1)}\beta_x + \frac{P(y=0|x)}{P(y=0)}(1 - \beta_x) \qquad (3)$$

where $\beta_x := P(D = 1|x)$. Thus, the convex combination (3) is maximized at each $x$ if and only if we place probability 1 to the largest term. That is to say, an optimal method assigns the positive class 1 with probability 1 if the term $P(y=1|x)/P(y=1)$ is larger or assigns the negative class with probability 1 if $P(y=0|x)/P(y=0)$ is larger. In other words,

$$\beta_x := P(D = 1|x) = \begin{cases} 1, & if \ \frac{P(y = 1|x)}{P(y = 1)} > \frac{P(y = 0|x)}{P(y = 0)} \\ 0, & Otherwise \end{cases} \quad (4)$$

This is indeed the method proposed above in eq. (1).

**Q.E.D.**

Critically, we note that the optimal method of eq. (4), will not have the true $P(y = 1|x)$ at hand but an estimation $\hat{P}(y = 1|x)$ instead. Notice also that all the other quantities for the decision are known constants $P(y = 1)$ and $P(y = 0)$ (i.e., the thresholds). Therefore, the good performance of the method totally relies on having good estimates $\hat{P}(y = 1|x)$.

### 3.2 Threshold for maximizing macro F1-score

Unlike macro accuracy, there is no theoretical threshold for maximizing the F1-score [24]. In general, increasing the threshold results in a higher precision of the minority class at expense of its recall. It is, however, known that 0.5 is the upper bound on the optimal threshold for the F1-score [24]. In the absence of any additional information and tuning we set the threshold for the minority class to $(P(y = 1) + 0.5)/2$. The rationale behind this threshold is that it is set midway between the highest recall (training set prior) and the upper bound on optimal F1 (0.5) thresholds. It should be noted that this threshold is defined for binary class. The extension to multiclass will be considered in the future work.

## 4 Experimental Setup

Proper selection of base classifiers is essential as our method relies on having good posterior probability estimates. Previous work has shown that bagged probabilistic decision trees provide accurate posterior probability estimates [22]. These are, in fact, more reliable than that of other probabilistic classifiers such as logistic regression [22]. Additionally, decision trees are complete (if they grow deep enough), that is, they are able to approximate any function [34]. These reasons motivated our choice of unpruned J48 decision trees (implementation of C4.5 trees in Weka) [35] [29]. Although it is common to use Laplace smoothing for the leaf probabilities of the individual decision trees it can be detrimental when data are imbalanced [36]. Therefore, we did not use Laplace smoothing for the trees. In addition, we study ensembles with varying number of base classifiers; 5, 10, 15, 25, 50 and 100.

As a validation method, we run 5×2-fold cross-validation for each method on each data set. We use the Friedman test to find if there are differences in the performance of the methods and if the test is passed at 95% significance, a post-hoc Nemenyi test is performed to identify any pairwise differences [37]. We used the R statistical environment (http://www.r-project.org/) with corresponding packages and default parameters, unless otherwise specified.

### 4.1 Data sets

We used 36 imbalanced binary data sets (Table 2); 14 from the HDDT repository (http://www3.nd.edu/~dial/hddt/), 19 from the KEEL repository (http://sci2s.ugr.es/keel/imbalanced.php) and 3 from the UCI repository (https://archive.ics.uci.edu/ml/datasets.html).

**Table 2.** Overview of the binary data sets obtained from UCI, HDDT* and KEEL† repositories

| Dataset | #Inst | #Attr | #Num | %Min | Dataset | #Inst | #Attr | #Num | %Min |
|---|---|---|---|---|---|---|---|---|---|
| pima | 768 | 8 | 8 | 0.345 | br-y† | 277 | 9 | 0 | 0.292 |
| ion | 351 | 34 | 34 | 0.359 | cl0vs4† | 173 | 13 | 13 | 0.075 |
| sonar | 208 | 60 | 60 | 0.466 | ecoli4† | 336 | 7 | 7 | 0.059 |
| spectf* | 267 | 44 | 44 | 0.206 | hab† | 306 | 3 | 3 | 0.265 |
| phon* | 5404 | 5 | 5 | 0.293 | led7_xvs1† | 443 | 7 | 7 | 0.083 |
| page* | 5473 | 10 | 10 | 0.102 | pb-1-3vs4† | 472 | 10 | 10 | 0.059 |
| ism* | 11180 | 6 | 6 | 0.023 | shut-0vs4† | 1829 | 9 | 9 | 0.067 |
| letter* | 20000 | 16 | 16 | 0.039 | vow0† | 988 | 13 | 13 | 0.091 |
| satim* | 6430 | 36 | 36 | 0.097 | yst-2vs4† | 514 | 8 | 8 | 0.099 |
| compu* | 13657 | 20 | 20 | 0.038 | yst4† | 1484 | 8 | 8 | 0.034 |
| segm* | 2310 | 19 | 19 | 0.143 | glass6† | 214 | 9 | 9 | 0.135 |
| oil* | 937 | 49 | 49 | 0.044 | new-th1† | 215 | 5 | 5 | 0.163 |
| estate* | 5322 | 12 | 12 | 0.12 | wisc† | 683 | 9 | 9 | 0.345 |
| hypo* | 2000 | 24 | 6 | 0.061 | car-gd† | 1728 | 6 | 0 | 0.04 |
| boun* | 3505 | 175 | 0 | 0.035 | flare-F† | 1066 | 11 | 0 | 0.04 |
| cred* | 1000 | 20 | 7 | 0.3 | kdd† | 1642 | 41 | 26 | 0.032 |
| hrt-v* | 133 | 9 | 4 | 0.233 | veh0† | 846 | 18 | 18 | 0.235 |
| ab9-18† | 731 | 8 | 7 | 0.057 | w-red-4† | 1599 | 11 | 11 | 0.033 |

**Table 3.** Overview of the multiclass data sets, all from the KEEL repository

| Dataset | #Inst | #Attr | #Num | %Min | #Class |
|---|---|---|---|---|---|
| contraceptive | 1473 | 9 | 6 | 0.226 | 3 |
| dermatology | 366 | 34 | 34 | 0.0546 | 6 |
| balance | 625 | 4 | 4 | 0.0784 | 3 |
| penbased | 1100 | 16 | 16 | 0.0954 | 10 |
| shuttle | 2175 | 9 | 9 | 0.00092 | 5 |
| wine | 178 | 13 | 13 | 0.2696 | 3 |
| yeast | 1484 | 8 | 8 | 0.00337 | 10 |
| pageblocks | 548 | 10 | 10 | 0.00547 | 5 |
| thyroid | 720 | 21 | 21 | 0.0236 | 3 |
| ecoli | 336 | 7 | 7 | 0.0059 | 8 |
| autos | 159 | 25 | 15 | 0.01887 | 6 |
| glass | 214 | 9 | 9 | 0.04206 | 6 |
| new-thyroid | 215 | 5 | 5 | 0.1395 | 3 |
| hayes-roth | 132 | 4 | 4 | 0.2273 | 3 |
| lymphography | 148 | 18 | 3 | 0.0135 | 4 |

Only the complete instances of data were used and any constant attributes were removed. Table 2 shows summary of the data sets of our experiment. The attributes of the data sets are a mix of numerical, nominal and numerical-nominal mixed. For multiclass setting we used 15 data sets from the KEEL repository (Table 3).

### 4.2    Methods and evaluation

For binary class, we included EB-bagging as the baseline method along with RB-bagging and SMOTE-bagging which are state-of-the-art methods (see Sect. 2.4 for details). The SMOTE parameters were set to 500% oversampling of the minority class and 120% undersampling of the majority, resulting in a balanced training set.
For completeness, we briefly compare with Platt scaling [38] on binary data sets. Platt scaling is a well-known calibration method that transforms output scores into probability estimates. Platt scaling fits a logistic regression model to the continuous classifiers outputs with the class labels as the dependent variable. To avoid over-fitting, as in the original paper [38], we fitted the logistic regression to the data from 3-fold cross-validation with transformed class labels as the dependent variable. A decision threshold of 0.5 was then applied to the calibrated probabilities to get the class labels. We deemed its inclusion as relevant since, as our method, Platt scaling is an easy-to-apply *a posteriori* probability "correction" over a classifier's output.

We evaluate the methods on three performance measures; area under the PR curve (AUCPR), macro accuracy and macro F1-score (see Sect. 2.1 for details). Note that AUCPR is computed using the posterior probability estimates while the other two measures make use of the class labels. As all the methods tested here use a threshold to obtain the class labels it is important to use proper thresholds. Rebalancing based ensembles commonly use the threshold of 0.5. This might not be appropriate for imbalanced data sets. Therefore, we also evaluate the methods on their "full potential". Full potential of a method is defined as the highest measure achieved over all possible thresholds. We approximate the full potential by a simple grid search over the thresholds (from 0 to 1 in steps of 0.01). Finally, we also use the stratified Brier score (mean squared error) to evaluate the calibration of the posterior probabilities. The Brier score for a class is calculated as the average squared difference between the estimated probability and the ideal probability.

## 5    Results and discussion

In this section, we compare the performance of the proposed method with other methods and provide further empirical insights.

### 5.1    Binary data sets

We will first investigate the effects of ensemble size. Fig. 1shows performance of all the methods with varying ensemble sizes. We note that, not surprisingly, area under the ROC curve showed a much more cluttered picture which we omit in the interest of space limitations.

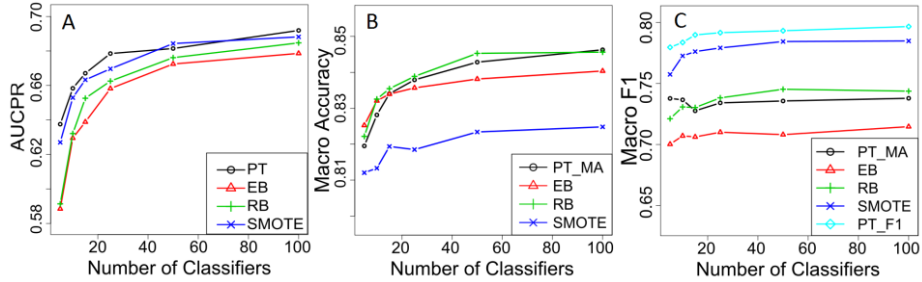

**Fig. 1.** Average test performance across datasets for different number of classifiers; (A) Area under the PR curve, (B) Macro accuracy and (C) Macro F1. The interpolated lines are shown for convenience.

The following general observations can be made: (1) all methods improve with ensemble size for all the measures, (2) RB-, PT- and EB-Bagging perform better on the macro accuracy while SMOTE-Bagging performs better on the macro F1 score. This again shows that in general different methods are suitable for different performance measures and (3) as expected, EB-Bagging looks like a poor version of RB-Bagging.

### 5.1.1 *Area under the Precision-Recall curve.*

As it can be readily observed PT-Bagging shows on average overall better performance on the area under the PR curve for all ensemble sizes followed by SMOTE-Bagging (Fig.1 A). This indicates that PT-Bagging is able to, in general, achieve comparatively larger means of precision and recall (F-measures). The Friedman test revealed a significant difference between the methods ($P<0.05$). The post-hoc pairwise tests showed a significant difference between PT-Bagging and EB-Bagging. There is some evidence that probability estimation of decision trees might deteriorate when categorical attributes are present [39]. Therefore, we repeated the tests for the 27 (out of 36) data sets with only numerical attributes. As before, this revealed a significant difference between the methods with significant difference between PT-Bagging and EB-Bagging. We also found a near significant difference between PT-Bagging and RB-Bagging ($P\sim0.055$).

### 5.1.2 *Macro accuracy*

We then applied the prior probabilities as threshold to maximize macro accuracy which we call $PT_{MA}$-bagging. Fig.1 B shows that while resampling methods offer similar or higher performance as $PT_{MA}$-bagging when a small number of classifiers are employed, as shown by Maloof [19]. However, the performance gain is larger for $PT_{MA}$-bagging as more classifiers are added, indicating that once variance is reduced through the averaged prediction, the error that is left comes mostly from bias. Thus, this result suggests that the bias of $PT_{MA}$-bagging is lower than that of resampling methods.

In order to gain insight on the bias of the methods we tested the null hypothesis that the difference between the recall of the two classes, i.e., two parts of the macro accuracy, is equal to zero. The null hypothesis was rejected for all the methods except $PT_{MA}$-bagging (single sample t-test: PT-Bagging $t$=-1.235, $P$=0.225; all other methods $P<0.05$) for ensembles with 100 classifiers. This suggests that $PT_{MA}$-bagging is not biased towards either of the classes (Fig. 2). Furthermore, only EB-Bagging shows higher recall for the minority class, suggesting a possible rebalancing overcompensation for the minority class.

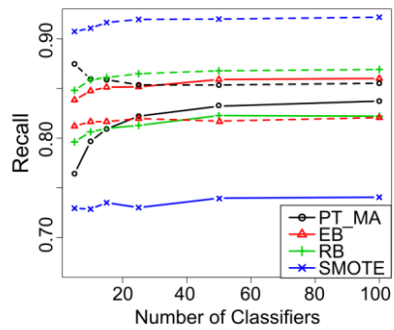

**Fig. 2.** Recall separated for the minority class (solid line) and the majority class (dashed line).

On average $PT_{MA}$-bagging showed the highest number of wins and fewer losses compared to other methods (Table 4, left). Friedman test revealed significant difference between the methods. The post-hoc test revealed significant difference between the following pairs of methods: $PT_{MA}$-bagging, EB-bagging; PT-bagging, SMOTE-bagging and RB-bagging, SMOTE-bagging (all $P<0.05$). There was close to trend level difference between RB-bagging and EB-bagging ($P\sim0.102$). Tests on data sets with only numerical attributes showed the same significant differences.

**Table 4.** Win/Tie/Loss table of macro accuracy (left) and macro F1-measure (right). The method of the row is compared against the method of the column where wins/ ties/losses.

|  | EB | RB | SMOTE |  | EB | RB | SMOTE |
|---|---|---|---|---|---|---|---|
| $PT_{MA}$ | 26/1/9 | 20/1/15 | 25/0/11 | $PT_{F1}$ | 34/0/2 | 27/1/8 | 18/0/18 |
| EB | - | 9/4/23 | 21/1/14 | EB | - | 1/2/33 | 3/1/32 |
| RB | - | - | 24/2/10 | RB | - | - | 4/1/31 |

A Friedman test on the full potential macro accuracy revealed significant differences between methods ($P<0.0014$) and post-hoc comparisons revealed both $PT_{MA}$-bagging ($P\sim0.03$) and SMOTE-bagging ($P\sim0.03$) to have a significantly higher potential than EB-bagging. RB-bagging did not show significantly different potential than EB-bagging ($P>0.26$). Other pairwise differences were not significant.

We then analyzed how much subpar each method performs compared to its full potential. The average difference between the maximum possible accuracy and the method accuracy was calculated across cross-validation folds and datasets. The results suggest that PT-Bagging performs closest to the best possible solution (Table 5 left).

**Table 5.** Mean difference between performed macro accuracy and maximum attainable macro accuracy (expressed in %) on the left, and between performed macro F1 and maximum attainable macro F1 on the right..

| $PT_{MA}$ | EB | RB | SMOTE | $PT_{F1}$ | EB | RB | SMOTE |
|---|---|---|---|---|---|---|---|
| 0.72% | 0.99% | 0.86% | 3.16% | 0.91% | 7.84% | 5.14% | 1.45% |

### 5.1.3 *F1 score and plug-in potency*

An important benefit of our method is that a learned ensemble can be used to make predictions that optimize any measure of interest by applying appropriate threshold (or misclassification cost) *a posteriori*, i.e., by thresholding differently. We *a posteriori* applied the threshold described in Sect. 3.2 to the same ensembles used for $PT_{MA}$-bagging resulting in $PT_{F1}$-bagging (Fig. 1C). The results are shown in Fig. 3 and Table 5. Note that methods have been proposed to estimate the optimal threshold for the F1 measure but they require tuning data and are amenable to the Winner's curse [24]. Therefore, using such methods is difficult for most of the datasets used here as they are relatively small. Also our aim here was to test a tuning free threshold. Finding better thresholds can conceivably further improve the F1 score.

$PT_{F1}$-bagging significantly improved macro F1 over the $PT_{MA}$-bagging, EB-bagging and RB-bagging (Friedman test $P<2.12e-12$, all post-hoc pairwise tests $P<0.05$, see Fig. 3C). $PT_{F1}$-bagging was not significantly different than SMOTE-bagging. This result supports our claim that PT-bagging passes as a plug-in method where the threshold can be set *a posteriori* according to the performance measure of interest.

Similar to macro accuracy, we analyzed the full potential of the methods for macro F1-score (Table 5). The Friedman test was nearly significant ($P\sim0.053$) and pairwise post-hoc tests only revealed trend level difference between SMOTE-bagging and EB-bagging ($P\sim0.066$). This suggests that all methods have the potential to achieve similarly high macro F1 if appropriate thresholds could be identified. However, we compared the macro F1 performance of methods against the maximum attainable macro F1, and we found that $PT_{F1}$ is particularly well-calibrated for the macro F1-measure as Table 5 (right) indicates.

### 5.1.4 *Posterior probability calibration*

We have shown that PT-bagging performs competitively on different measures, which is possible only with well calibrated posterior probabilities. In the following we argue that PT-bagging estimates well calibrated posterior probabilities. An empirical study by Niculescu-Mizil and Caruana showed that, bagged decision trees and neural networks predict well calibrated probabilities, making *a posteriori* calibration -with e.g., Platt scaling- unnecessary [40]. However, calibration is a relatively understudied problem for imbalanced data. In this direction, a recent study proposes to correct the calibration for undersampling [15] and another study proposes use of undersampling to obtain calibrated probabilities [41]. Those studies use the (stratified) Brier score to quantify the calibration. Wallace and Dahabreh [42] find that undersampling with bagging leads to lower Brier score, i.e. better calibration, for the minority class without sacrificing the overall score. We find similar results in our experiments, especially we find that the rebalancing methods have a significantly lower Brier score for the minority class while PT-bagging has a significantly lower Brier score for the majority class (both cases Friedman test: $P<2.2e-16$ and all pairwise post-hoc Nemenyi tests $P<0.05$ except for the RB-bagging, SMOTE-bagging pair).

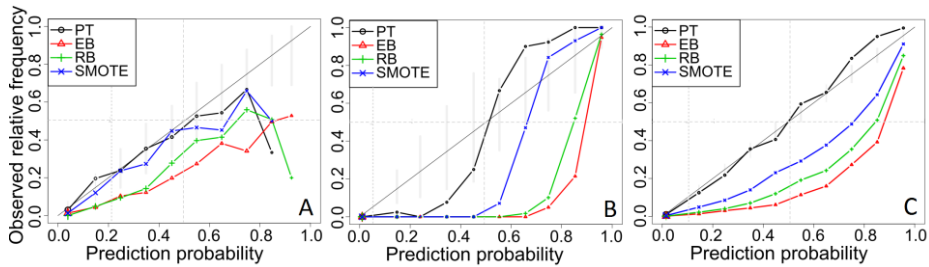

**Fig. 3.** Reliability plots for 3 datasets (A) spectf (UCI), (B) pb-1-3vs4 (KEEL) and (C) satim (HDDT). We used 10 bins to discretize the posterior probability for the minority class (x-axis) for all the runs and folds and then the corresponding observed frequencies of the minority class (y-axis) were calculated for each bin. A method lining up with the diagonal is well calibrated while below the diagonal is overestimating the probability. The left vertical dashed line denotes the prior of the minority class.

This seemingly negative result can be attributed to a shortcoming of the Brier score. The Brier score prefers crisp posteriors, e.g. the Brier score for the minority class will become zero if all the posteriors were 1. Even though this might seem like a good thing it is not necessarily always the case. Intuitively, any threshold will have little or no effect on the class labels if the posteriors are crisp (as a consequence the plug-in paradigm will not work). The information in the non-crisp posteriors is ignored by the Brier score and overestimated probabilities will lead to a lower Brier score. Besides having only crisp posteriors is unrealistic for most of the real world data. Developing new measures to quantify calibration is out of scope of this paper, so we resort to visualization using reliability plots.

Here we show three reliability plots as exemplars (Fig. 3). In general, PT-bagging probabilities were well calibrated (close to the diagonal) for majority of the data sets (21 out of 36, see supplementary material) while all other methods seem to overestimate the posteriors of the minority class (hence lower Brier scores). Moreover, then PT-bagging failed to estimate the posteriors correctly, i.e. they were not aligned with the diagonal, other methods failed too. Our analysis suggests that PT-bagging can estimate well-calibrated posteriors. Therefore, the (stratified) Brier score can be a misleading measure and should be used with caution.

*5.1.5*   ***Comparison with Platt scaling***

We applied Platt scaling with a decision threshold of 0.5 to the probabilities of the same 100 classifiers as PT-bagging on all the 36 binary class datasets. We compare the results with the two well performing methods. $PT_{MA}$-bagging outperformed Platt scaling on macro-accuracy (Paired Wilcoxon test P<1e-8, Win/Tie/Loss=34/0/2). Platt scaling performed relatively better on F1-score than macro accuracy. Nevertheless, SMOTE-bagging still outperformed Platt (Paired Wilcoxon test P~0.021, Win/Tie/Loss=21/0/15). In conclusion, probability calibration using Platt scaling did not provide improvement over the methods compared here.

**5.2   Multiclass data sets**

Another advantage of our method is that it can be directly extended to handle multiclass data. For multiclass data the thresholding can be applied by dividing the posteriors by appropriate threshold probabilities, for instance to maximize macro accuracy the thresholds equal prior probabilities of the respective class (see Algorithm 1). We evaluated this approach on 15 multiclass datasets (Table 3). For comparison we used the UnderBagging to OverBagging method (UnderOver) proposed by Wang and Yao [43]. Their method under- or oversamples instances of different classes in proportion to the majority class size controlled by a parameter called *a*. We varied this parameter in 10, 25, 50 and 100. We also added another method in which instances from all the classes are undersampled to match the smallest class, a similar sampling mechanism as EB.

We used similar settings to the ones we used for the binary data sets, i.e., simple bootstrap sampling, 5×2-fold cross-validation with unpruned J48 decision trees. We used only single ensemble size of 100. As there were too many methods and not many data sets it is unlikely that statistical tests will reveal any significance. Therefore, we selected a single method from all the competitors that had maximum average macro accuracy. This method with average accuracy of 0.756 was UnderOver-bagging with the parameter *a* set to 50. Wilcoxon paired signed rank test between this method and $PT_{MA}$-bagging revealed trend level significant difference ($P$~0.0946) with $PT_{MA}$-bagging showing higher average macro accuracy (0.789). Furthermore, of a total of 15 multiclass data sets, $PT_{MA}$-bagging had 8 overall wins against all other methods, while the second best method with 4 wins was UnderOver-bagging (sampling with *a*=10). This result shows that PT-bagging can be employed for multiclass data sets when appropriate thresholds are available, though further tests are needed to confirm statistical significance.

## 6 Conclusions and Future Work

We proposed a simple plug-in method PT-bagging for imbalanced data classification. Our method relies on simple bootstrap sampling, which preserves the natural class distribution, to create a bagging ensemble followed by thresholding to assign class labels. Our results and analyses show that PT-bagging performs competitively compared to state-of-the-art rebalancing methods. Furthermore, it does so for two different performance measures, macro accuracy and macro F1-score, by adjusting the threshold *a posteriori*. In addition, we analyzed the results providing several insights, especially we found that; (1) PT-bagging is less biased towards any of the classes, (2) it performs close to its full potential and (3) a potential shortcoming in using the Brier score for quantifying probability calibration. We also demonstrated that PT-bagging can be directly extended to multiclass data when appropriate thresholds are available. Taken together, our results and analyses open up new possibilities and a basis for development of new threshold-moving methods.

An important, but understudied, question is whether to use the *natural* class distribution for learning [20]. Weiss and Provost [44] studied this question empirically and concluded that generally a different than natural class distribution lead to better performance. Our results stand in contrast with their conclusion. An important difference between theirs and our method -which can partly explain the different conclusions- is the use of a single classifier (C4.5) versus an ensemble of decision trees. As we have shown simple bootstrap performs poorly with small ensembles and the performance improves with the ensemble size. Taking this into consideration, we conclude that bagged decision trees with enough classifiers can model the data with natural class distribution.

There are several possible directions we can take in the future. One promising direction is the investigation of use of data properties to improve the sampling mechanism (see e.g., [7]). Such methods can leverage the intrinsic properties of the data and help with difficult situations, e.g., small disjuncts in the minority class. We would like to investigate suitability of our method in environments with dynamic costs as our method avoids the computationally expensive re-training.

**Acknowledgements**
We thank Dr. Harikrishna Narasimhan, Prof. Jesse Davis and Prof. Hendrik Blockeel for fruitful discussions. We also thank three anonymous ECML-PKDD 2016 reviewers that provided insightful comments. KRP was partly supported by the WT-MIT fellowship 103811AI.

## 7 References

1. Yang, Q., Wu, X.: 10 challenging problems in data mining research. International Journal of Information Technology & Decision Making. 597–604 (2006).
2. Yang, Z., Tang, W.: Association rule mining-based dissolved gas analysis for fault diagnosis of power transformers. IEEE Transactions on Systems, Man, and Cybernetics, Part C: Applications and Reviews. 39.6, 597–610 (2009).

3. Perdisci, R., Gu, G., Lee, W.: Using an ensemble of one-class svm classifiers to harden payload-based anomaly detection systems. IEEE Sixth International Conference on Data Mining (ICDM). (2006).
4. Fawcett, T., Provost, F.: Adaptive fraud detection. Data mining and knowledge discovery. (1997).
5. Mazurowski, M., Habas, P., Zurada, J.: Training neural network classifiers for medical decision making: The effects of imbalanced datasets on classification performance. Neural networks. 427–436 (2008).
6. Galar, M., Fernandez, A., Barrenechea, E., Bustince, H., Herrera, F.: A Review on Ensembles for the Class Imbalance Problem: Bagging-, Boosting-, and Hybrid-Based Approaches. IEEE Transactions on Systems, Man, and Cybernetics, Part C (Applications and Reviews). 42, 463–484 (2012).
7. He, H., Garcia, E.: Learning from imbalanced data. IEEE Transactions on Knowledge and Data Engineering,. 1263–1284 (2009).
8. Batista, G.E.A.P.A., Prati, R.C., Monard, M.C.: A study of the behavior of several methods for balancing machine learning training data. ACM SIGKDD Explorations Newsletter. 6, 20 (2004).
9. Nitesh V. Chawla, Kevin W. Bowyer, Lawrence O. Hall, W.P.K.: SMOTE: Synthetic Minority Over-sampling Technique. Journal of Artificial Intelligence Research. 16, 321–357 (2002).
10. Chawla, N., Bowyer, K.: SMOTE: synthetic minority over-sampling technique. Journal of artificial intelligence research. (2002).
11. Estabrooks, A., Jo, T., Japkowicz, N.: A multiple resampling method for learning from imbalanced data sets. Computational Intelligence. 20, 18–36 (2004).
12. Dietterich, T.: Ensemble methods in machine learning. Multiple classifier systems. (2000).
13. Hido, S., Kashima, H., Takahashi, Y.: Roughly balanced bagging for imbalanced data. Statistical Analysis and Data Mining. 2, 412–426 (2009).
14. Seiffert, C., Khoshgoftaar, T.M., Van Hulse, J., Napolitano, A.: RUSBoost: A hybrid approach to alleviating class imbalance. IEEE Transactions on Systems, Man, and Cybernetics Part A:Systems and Humans. 40, 185–197 (2010).
15. Pozzolo, A.D., Caelen, O., Johnson, R., Bontempi, G.: Calibrating Probability with Undersampling for Unbalanced Classification. 2015 IEEE Symposium on Computational Intelligence and Data Mining. 159–166.
16. Pozzolo, A.D., Caelen, O., Bontempi, G.: When is undersampling effective in unbalanced classification tasks? Machine Learning and Knowledge Discovery in Databases. (2015).
17. Chawla, N., Hall, L., Joshi, A.: Wrapper-based computation and evaluation of sampling methods for imbalanced datasets. Proceedings of the 1st international workshop on Utility-based data mining. ACM. (2005).
18. Wang, S., Yao, X.: Multiclass imbalance problems: Analysis and potential solutions. IEEE Transactions on Systems, Man, and Cybernetics, Part B: Cybernetics 42.4. 1119–1130 (2012).
19. Maloof, M.: Learning when data sets are imbalanced and when costs are unequal and unknown. CML-2003 workshop on learning from imbalanced data sets II. (2003).

20. Provost, F.: Machine learning from imbalanced datasets. In: Proc. of the AAAI'2000 Workshop on ImbalancedData Sets (2000).
21. Zhou, Z., Liu, X.: Training cost-sensitive neural networks with methods addressing the class imbalance problem. IEEE Transactions on Knowledge and Data Engineering 18.1. (2006).
22. Provost, F., Domingos, P.: Tree induction for probability-based ranking. Machine Learning. (2003).
23. Narasimhan, H., Vaish, R., Agarwal, S.: On the statistical consistency of plug-in classifiers for non-decomposable performance measures. Advances in Neural Information Processing Systems. (2014).
24. Lipton, Z., Elkan, C., Naryanaswamy, B.: Optimal thresholding of classifiers to maximize F1 measure. Machine Learning and Knowledge Discovery in Databases. (2014).
25. Elkan, C.: The foundations of cost-sensitive learning. International joint conference on artificial intelligence. (2001).
26. Kuncheva, L.: Combining pattern classifiers: methods and algorithms. (2004).
27. Breiman, L.: Bagging predictors. Machine Learning. 24, 123–140 (1996).
28. Bauer, E., Kohavi, R.: An empirical comparison of voting classification algorithms: Bagging, boosting, and variants. Machine learning. 105–139 (1999).
29. Hastie, T., Tibshirani, R., Friedman, J.: The elements of statistical learning. (2009).
30. Błaszczyński, J., Stefanowski, J., Idkowiak, Ł.: Extending bagging for imbalanced data. Proceedings of the 8th International Conference on Computer Recognition Systems CORES. (2013).
31. Drummond, C., Holte, R.: C4. 5, class imbalance, and cost sensitivity: why under-sampling beats over-sampling. Workshop on learning from imbalanced datasets II. (2003).
32. Sheng, V., Ling, C.: Thresholding for making classifiers cost-sensitive. Proceedings of the national conference on artificial intelligence AAAI. (2006).
33. Hernández-Orallo, J., Flach, P., Ferri, C.: A unified view of performance metrics: Translating threshold choice into expected classification loss. Journal of Machine Learning Research (JMLR). 13(Oct), 2813–2869. (2012).
34. Breiman, L.: Some infinity theory for predictor ensembles. (2000).
35. Pang, S., Gong, J.: C5.0 Classification Algorithm and Application on Individual Credit Evaluation of Banks. Systems Engineering - Theory & Practice. 29, 94–104 (2009).
36. Zadrozny, B., Elkan, C.: Learning and making decisions when costs and probabilities are both unknown. Proceedings of the seventh ACM SIGKDD international conference on Knowledge discovery and data mining. ACM. (2001).
37. Demšar, J.: Statistical comparisons of classifiers over multiple data sets. The Journal of Machine Learning Research (JMLR). 7(Jan), 1–30 (2006).
38. Platt, J.: Probabilistic outputs for support vector machines and comparisons to regularized likelihood methods. Advances in large margin classifiers. (1999).
39. Zhang, K., Fan, W., Buckles, B.: Discovering unrevealed properties of probability estimation trees: On algorithm selection and performance explanation. IEEE Sixth International Conference on Data Mining, ICDM. (2006).
40. Niculescu-Mizil, A., Caruana, R.: Predicting good probabilities with supervised

learning. In: Proceedings of the 22nd international conference on Machine learning (ICML) (2005).
41. Wallace, B., Dahabreh, I.: Improving class probability estimates for imbalanced data. Knowledge and Information Systems. 33–52 (2014).
42. Wallace, B., Dahabreh, I.: Class probability estimates are unreliable for imbalanced data (and how to fix them). IEEE 12th International Conference on Data Mining (ICDM). (2012).
43. Wang, S., Yao, X.: Diversity analysis on imbalanced data sets by using ensemble models. Computational Intelligence and Data Mining, 2009. CIDM'09. IEEE Symposium on. IEEE. (2009).
44. Weiss, G., Provost, F.: Learning when training data are costly: the effect of class distribution on tree induction. Journal of Artificial Intelligence Research. 315–354 (2003).